%% file: arxiv_paper.tex
\documentclass[11pt]{article}

\usepackage{times}
\usepackage{epsfig}
\usepackage{graphicx}
\usepackage{amsmath}
\usepackage{amssymb}

\usepackage{caption}
\usepackage{subcaption}
\usepackage{url}

\input{commands}

\newlength{\figwidth}
\usepackage{hyperref}
\hypersetup{
    unicode=false,          
    pdftoolbar=true,        
    pdfmenubar=true,        
    pdffitwindow=false,     
    pdfstartview={FitH},    
    pdfauthor={Angel Villar-Corrales and Veniamin I. Morgenshtern},     
    pdfcreator={Angel Villar-Corrales and Veniamin I. Morgenshtern},   
    pdfproducer={Producer}, 
    pdfnewwindow=true,      
    colorlinks=true,       
     linkcolor=blue!60!black,          
    citecolor=blue!60!black,        
    filecolor=blue!60!black,      
    urlcolor=blue!60!black,
     hypertexnames=false
}

\begin{document}
\title{Scattering Transform Based Image Clustering using \\ Projection onto Orthogonal Complement}

\pdfinfo{
/Title  (Scattering Transform Based Image Clustering using Projection onto Orthogonal Complement)
/Author (Angel Villar-Corrales)
/Keywords ()
}


\author{
	\parbox{\linewidth}{\centering
		Angel Villar-Corrales and Veniamin~I.~Morgenshtern\\\vspace{0.3cm}
		Chair of Multimedia Communications and Signal Processing,\\ University of Erlangen-Nuremberg,\\ 
		Erlangen 91058, Germany\\
		E-mail: angel.corrales.villar@fau.de, veniamin.morgenshtern@fau.de
	}
}

\maketitle

\begin{abstract}

	In the last few years, large improvements in image clustering have been driven by the recent advances in deep
	learning. However, due to the architectural complexity of deep neural networks, there is no mathematical theory that explains
	the success of deep clustering techniques.
	In this work we introduce Projected-Scattering Spectral Clustering (PSSC), a state-of-the-art, stable, and fast algorithm for image  clustering, 
	which is also mathematically interpretable. 
	PSSC includes a novel method to exploit the geometric structure of the scattering transform of small images.
	This method is inspired by the observation that, in the scattering transform domain, the subspaces formed by the eigenvectors corresponding to the few largest eigenvalues  of the data matrices of individual classes are nearly shared among different classes. 
	Therefore, projecting out those shared subspaces reduces the intra-class variability, substantially increasing the clustering performance.
	We call this method ‘Projection onto Orthogonal Complement’ (POC).
	Our experiments demonstrate that PSSC obtains the best results among all shallow clustering algorithms. Moreover,
	it achieves comparable clustering performance to that of recent state-of-the-art clustering techniques, while reducing the execution time by more than one order of magnitude. In the spirit of reproducible research, 
	we publish a high quality code repository along with the paper \footnote{\url{https://github.com/vmorgenshtern/scattering-clustering}}.

\end{abstract}

\begin{figure*}[t]
	\centering
	\includegraphics[width=0.9\linewidth]{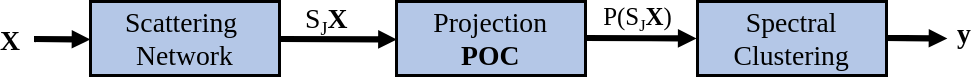}
	\caption{PSSC framework. Clustering is performed by cascading three well understood mathematical operators. First, the scattering transform of the images is computed. Second, intra-class variabilities are reduced using POC projection step. Third, the final cluster assignments are obtained by applying spectral clustering to the processed scattering representations.}
	\label{fig:clustering pipeline}
	\vspace{-0.15cm}
\end{figure*}

\section{Introduction}\label{sec:introduction}

Image clustering is a fundamental computer vision task that aims at grouping images into clusters according to certain similarity metric, so that similar images belong to the same cluster and the images in different clusters are dissimilar. Image clustering is used in several machine learning and computer vision applications including image retrieval~\cite{Chen_ContentBasedImageRetrievalClustering_2003, Murthy_ContetBasedImageRetrievalKMeansClustering_2010}, data mining~\cite{Verma_ComparativeStudyClusteringForDataMining_2012} and image segmentation~\cite{Chuang_FuzzyKMeansImageSegmentation_2006, Dhanachandra_ImageSegementationThroughClustering_2015}.

Due to large variance in shape, style and appearance, clustering real images is considered a challenging task. Traditional clustering methods, such as $k$-means \cite{Macqueen_KMeansClustering_1967}, mean-shift \cite{Comaniciu_MeanShift_2002}, DBSCAN \cite{Ester_DBSCANDensityBasedClustering_1996}, hierarchical clustering \cite{Johnson_HierarchicalClustering_1967} or Gaussian mixture-models \cite{Reynolds_GaussianMixtureModels_2009}, rely on computing the distance between handcrafted features extracted from the dataset samples. These techniques fail at clustering complex high-dimensional images, since engineered features often cannot capture the complexity of real data.

Spectral clustering \cite{Ng_SpectralClustering_2002, Luxburg_SpectralClusteringTutorial_2007} generalizes ideas from spectral graph theory to cluster samples using the eigenvectors of matrices derived from data. Despite often outperforming the aforementioned algorithms, spectral clustering also relies on engineered features and suffers from the same drawbacks as the traditional approaches. Traditional and spectral clustering methods are often denoted as \textit{shallow clustering} algorithms.

The recent advances in deep neural networks (DNNs) \cite{Lecun_CNNsVision_2010, Krizhevsky_Alexnet_2012, Zisserman_VGG_2014, He_DeepResidualLearningResNet_2016} have significantly improved the state-of-the-art for image clustering, giving birth to a new family of clustering algorithms commonly known as \textit{deep clustering}. These methods perform (separately or jointly) two different processes: image features are learned from data using DNNs, and then these are used to group the images into clusters. By minimizing a clustering loss function~(e.g., $k$-means loss~\cite{Yang_TowrdsKMeansFriendlySpacesDeepClustering_2017} or cluster hardening loss \cite{Xie_UnsupervisedDeepEmbeddingClustering_2016}), the model is able to learn feature representations optimized for clustering.

Several architectures have been employed for deep clustering, including CNNs~\cite{Caron_DeepClusteringUnsupervisedLearningVisualFeatures_2018, Yang_JointUnsupervisedLearningImageClusters_2016joint}, autoencoders~\cite{Mcconville_DeepClusteringManifoldAutoencodedEmbedding_2019, Guo_DeepClusteringConvolutionalAutoencoders_2017, Ren_DeepDensityBasedClustering_2020, Yang_TowrdsKMeansFriendlySpacesDeepClustering_2017} or GANs~\cite{Donahue_AdversarialFeatureLearning_2016, Mrabah_AdversarialDeepEmbeddedClustering_2020}. Despite their impressive performance for image clustering, the architecture complexity of DNNs, a deep cascade of learned modules and nonlinear operators \cite{Lecun_CNNsVision_2010}, makes them hard to analyze mathematically \cite{Szegedy_IntriguingPropertiesNeuralNetworks_2013, Mallat_UnderstandingDeepNetworks_2016, Zhang_RethinkingGeneralization_2016}. Furthermore, deep clustering techniques require several minutes to process mid-sized datasets, hence being highly computationally inefficient.

In this work, we propose Projected-Scattering Spectral Clustering (PSSC), an image clustering framework that obtains results comparable to those of state-of-the-art DNNs, while being efficient, robust and interpretable. Our clustering framework (depicted in \Figure{fig:clustering pipeline}) is composed of three steps. First, we process the images using a scattering network (ScatNet)~\cite{Bruna_InvariantScattering_2012}. This step generates image features that are invariant to small translations and linearizes intra-class variabilities. Second, we introduce a new processing technique, denoted \textit{Projection onto Orthogonal Complement} (POC), motivated by the geometrical structure of scattering features.
This method is inspired by the observation that, in the scattering transform domain, the subspaces formed by the eigenvectors corresponding to the few largest eigenvalues  of the data matrices of individual classes are nearly shared among different classes. 
Therefore, projecting out those shared subspaces reduces the intra-class variability, substantially increasing the clustering performance.
Finally, we cluster the processed feature vectors using an efficient and scalable spectral clustering algorithm inspired by~\cite{Huang_UltraScalableSpectralClustering_2019}.

In summary, the contributions of our work are as follows:

\begin{itemize}
	\item We introduce PSSC, a three-step framework to efficiently compute image clusters from an unlabeled image dataset. Unlike deep clustering techniques, our proposed pipeline is provably stable and mathematically tractable.
	
	\item We propose POC, a simple, but effective, geometrically motivated algorithm for processing scattering representations. We empirically demonstrate that the POC step simplifies deformations and structural variabilities, hence improving the clustering performance.
	
	\item Our experimental results show that the proposed framework outperforms all shallow clustering algorithms and obtains results comparable to those of state-of-the-art deep clustering models for four different benchmark datasets, while reducing execution runtime by more than one order of magnitude.

\end{itemize}

\section{Related Work}\label{sec:related}

\subsection{Scattering Transform}\label{sec:scattering}

ScatNet \cite{Mallat_GroupInvariantScattering_2012, Bruna_InvariantScattering_2012} is a multi-layer hierarchical network in which the convolutional kernels correspond to wavelet filters. Scattering transform is an invariant, stable, and informative signal representation obtained by cascading wavelet modulus decompositions followed by a lowpass filter. Below we review the basics of the scattering transform and we refer to \cite{Bruna_InvariantScattering_2012} for further details.

Similarly to the wavelet transform, the scattering transform starts with a \textit{mother wavelet} $\MotherWavelet(u)$. In the case of images, the wavelet basis is obtained by scaling $\MotherWavelet(u)$ and rotating it using a rotation group $\RotationGroup$ of $\mathbb{R}^2$. Therefore, for a certain scale $ j \in \mathbb{Z}$ and rotation $r \in \RotationGroup$, the wavelet function is defined as:

\begin{align}
& \MotherWavelet_{2^{j} r}(u) = 2^{2j} \MotherWavelet(2^{j} r^{-1} u ).
\end{align}

To simplify notation, we define $\Orientation = 2^{j} r$, with $|\Orientation|=2^{j}$. Each rotated and dilated wavelet extracts the energy of a signal located at a scale and orientation given by $\Orientation$.

Translation \textit{invariance} of a representation is a desirable property for computer vision tasks. However, due to the convolution operator, the wavelet transform is inherently translation \textit{covariant}. A translation invariant representation can be built by applying a non-linear operator that computes an informative average value of the wavelet coefficients.
Namely, scattering transform creates a locally translation invariant representation by computing the complex modulus non-linearity and averaging the results using a low-pass filter~$\ScalingFunction$. 

Given an image $\Input(u)$, where $u \in \mathbb{R}^2$ indexes pixel coordinates, the zero-order scattering coefficients $\ScatteringCoefficients[0]$ are obtained by averaging the signal energy using the low-pass filter:

\begin{align}
& \ScatteringCoefficients[0] \Input(u) = (\Input \star \ScalingFunction) (u) = \int \Input(v) \cdot \ScalingFunction(u-v) dv.
\end{align}

Higher-order translation invariant scattering coefficients can be computed by cascading wavelet transforms with modulus operators. Let $\Path = (\Orientation_1, \Orientation_2, ...,  \Orientation_m)$ denote an ordered sequence of wavelets, also referred as path. Then, the scattering coefficients for a path $\Path$ are computed as follows:

\begin{align}
& \ScatteringCoefficients[p] \Input(u) = | | |\Input \star \MotherWavelet_{\Orientation_1}|
\star \MotherWavelet_{\Orientation_2} | ... | \star \MotherWavelet_{\Orientation_m} | \star \ScalingFunction(u).    	
\end{align}

Since the architecture of the scattering transform is structured as the cascade of convolution and non-linear operators, it shares many similarities with the architecture of a CNN.
The difference is that the convolutional kernels are given by wavelets, which do not need to be learned.

Scattering features are locally translation invariant and stable to deformations~\cite{Anden_DeepScatteringSpectrum_2014}, hence they are useful unsupervised features for several signal processing and machine learning tasks. 

In their seminal work, Bruna and Mallat~\cite{Bruna_InvariantScattering_2012} used scattering representations as input features for support vector machines and PCA-based classifiers to obtain state-of-the-art results for classification of handwritten digits and textures in a \emph{supervised} learning setting.  

In \cite{Oyallon_ScatteringHybridNetworks_2018}, Oyallon \EtAl.~showed how hybrid architectures, obtained by combining the scattering transform with CNNs, outperform end-to-end classifiers for datasets composed of small images such as CIFAR-10 or STL-10.

More recently, Zarka \EtAl.~\cite{Zarka_ScatteringHomotopyDictionaryLearningClassification_2019} introduced a classification pipeline constructed by cascading well understood mathematical operators. Namely, the pipeline is composed of a ScatNet followed by dimensionality reduction, dictionary learning and a multi-layer perceptron (MLP) classifier. Using only predefined scattering representations, the authors reached higher accuracy than AlexNet~\cite{Krizhevsky_Alexnet_2012} on the ImageNet image classification challenge. 

Loosely inspired by the previous work, our method also profits from the reduced variability and invariances provided by the scattering transform. However, in contrast to those methods, we use the scattering representations to address an \emph{unsupervised} learning problem and we apply a novel processing step to make the feature representation more suitable for the task of clustering.

\subsection{Spectral Clustering}\label{sec:spectral clustering}

Spectral clustering \cite{Ng_SpectralClustering_2002, Luxburg_SpectralClusteringTutorial_2007, Stella_MulticlassSpectralClustering_2003, Zelnik_SelfTuningSpectralClustering_2005} refers to a family of clustering algorithms that use the top eigenvectors of a matrix computed from pairwise distances between samples. These algorithms have been widely used in practice due to their ability to 
deal with non-linearly separable datasets. For more details about the spectral clustering algorithms, we refer to~\cite{Luxburg_SpectralClusteringTutorial_2007}.

Despite its desirable clustering properties, spectral clustering algorithms do not gently scale to large datasets due to a computational complexity of $O(N^3)$, where $N$ is the number of samples in the dataset. This complexity makes it unfeasible to apply spectral clustering to datasets containing over a few thousand samples. To address this issue, novel \emph{approximate} clustering algorithms~\cite{Huang_UltraScalableSpectralClustering_2019, Yan_FastApproximateSpectralClustering_2009, He_FastSpectralClusteringExplicitFeatureMapping_2018} have been developed in order to reduce the computational complexity of spectral clustering so as to scale to arbitrarily large datasets.

\section{Projected-Scattering Spectral Clustering}\label{sec:Scattering Projected Spectral Clustering}

This section presents a novel clustering framework: PSSC. Let $\ImageDataset = \{\Image_i\}_{i=1}^N$ denote a dataset with $N$ images. PSSC aims at grouping these images into $\NumClusters$ disjoint clusters based on an intermediate image representation obtained through the scattering transform and a geometrically motivated projection step denoted POC. The resulting clustering pipeline is illustrated in \Figure{fig:clustering pipeline}.

\Section{sec:Scattering Projected Spectral Clustering} is organized as follows. In \Section{sec:Geometry of Scattering Transforms}, we investigate the spectral distribution of the scattering transform of small images, with special focus on correlation between features and eigenvalue distribution. In \Section{sec:Projection onto Orthogonal Complement}, we introduce the POC algorithm, which exploits the geometrical structure of scattering representations to reduce intra-class variability. Finally, in \Section{sec:Clustering Approach}, we introduce the clustering algorithm, inspired by \cite{Huang_UltraScalableSpectralClustering_2019}, which we use to compute the cluster assignments.

\begin{figure}[b]	
	\hspace{0.05\textwidth}
	\begin{subfigure}{0.3\textwidth}
		\hspace*{0\textwidth}
		\includegraphics[height=0.2\textheight]{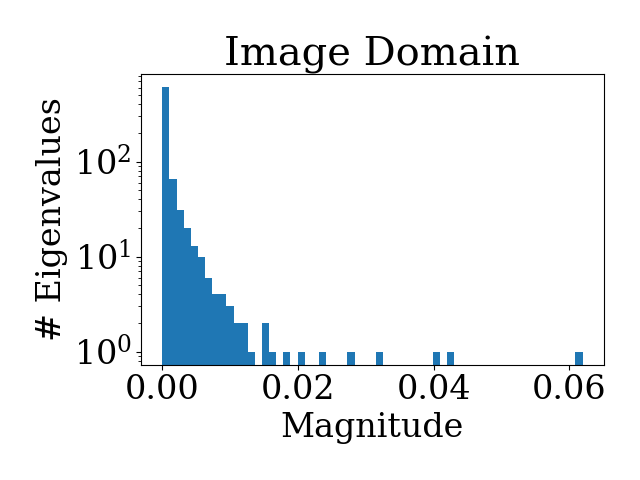}
		\label{fig:subfig1}
	\end{subfigure}
	\hspace{0.15\textwidth}
	\begin{subfigure}{0.3\textwidth}
		\centering
		\includegraphics[height=0.2\textheight]{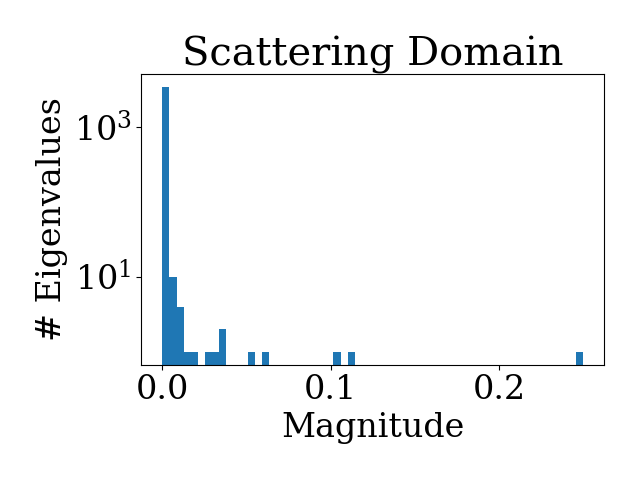}
		\label{fig:subfig2}
	\end{subfigure}
	\hspace{0.1\textwidth}
	\vspace{-0.7cm}
	\caption{Eigenvalue distribution of the covariance matrix of the MNIST test set in the image domain (left) and scattering domain (right). Eigenvalues have been normalized with respect to $\ell_1$ norm. In the scattering transform domain, the four largest eigenvalues explain over 50\% of the data variance, whereas close to 50 eigenvalues are needed in the image domain to obtain the same percentage.}
	\label{fig:eigenvalues mnist dataset}
\end{figure}

\subsection{Geometry of Scattering Transforms}\label{sec:Geometry of Scattering Transforms}

The scattering transform is the first step of our clustering framework, as depicted in \Figure{fig:clustering pipeline}. This step aims at transforming the input images into a more suitable representation for the task of clustering: $\ImageDataset \rightarrow \ScatteringFeatures$.

Invariance to small translations and rotations, as well as robustness to deformations, provided by the scattering transform representation, are desirable properties for pattern recognition tasks. Nevertheless, methods that rely on engineered input features are often outperformed by learned representations such as those obtained using DNNs.

In order to further improve the data representation provided by the scattering transform for our particular task of image clustering, we exploit the geometrical structure of the transformed data. In particular, we aim at minimizing the high correlations between scattering transform coefficients.

As found out by Bruna and Mallat in~\cite{Bruna_InvariantScattering_2012}, natural images have scattering coefficients that are strongly correlated across paths. 
\Figure{fig:eigenvalues mnist dataset} displays the eigenvalue distribution of the covariance matrix of the MNIST~\cite{LeCun_MnistDataset_1998} test-set in the image domain on the left and in the scattering domain on the right. The eigenvalues have been normalized with respect to the $\ell_1$ norm, so that they sum up to one. In the scattering domain, the four largest eigenvalues explain more that 50\% of the total variance. In fact, more that 95\% of the eigenvalues have an almost-zero magnitude.

This quick variance decay in the scattering domain indicates a strong correlation and redundancy between features.

\begin{figure}[b!]
	\centering
\begin{minipage}{.7\linewidth}
\begin{algorithm}[H]
	\caption{Principal angles between two subspaces}
	\label{algorithm:principal angles algorithm}
	\begin{algorithmic}
		\Procedure{Principal Angles Algorithm}{}
		\BState \emph{Inputs}:
		\State $\mathbf{U} = [\mathbf{u}_1, ..., \mathbf{u}_N] 
		\gets \textit{vectors forming subspace 1}$
		\State $\mathbf{V} = [\mathbf{v}_1, ..., \mathbf{v}_N]  
		\gets \textit{vectors forming subspace 2}$
		\State $D \gets \textit{number of principal angles to compute}$
		\BState \emph{Returns}:
		\State $\PrincipalAngles \gets \textit{principal angles between subspaces } \mathbf{U} \textit{ and } \mathbf{V}$
		\BState \emph{Algorithm}:
		\State $\mathbf{U^{\perp}} \gets [~]$
		\State $\mathbf{V^{\perp}} \gets [~]$
		
		\For{ $i$ in range [1, $D$]}
		
		\vspace{-0.7cm}
		\State \begin{align}
		\PrincipalAngle{i} = \min\limits_{\mathbf{u} \in \mathbf{U}, \mathbf{v} \in \mathbf{V}} \ &\arccos \left(\frac{\langle\mathbf{u}, \mathbf{v}\rangle}{||\mathbf{u}||_2||\mathbf{v}||_2} \right) \notag \\
		\text{s.t. } &\mathbf{u}^T \mathbf{u}^{\perp} = 0 \quad \text{for all} \quad \mathbf{u}^{\perp} \in \mathbf{U^{\perp}}  \notag \\
		&\mathbf{v}^T \mathbf{v}^{\perp} = 0  \quad \text{for all} \quad \mathbf{v}^{\perp} \in \mathbf{V^{\perp}}  \notag
		\end{align}
		\vspace{-0.6cm}
		
		\State $\mathbf{U^{\perp}} \gets [\mathbf{U^{\perp}}, \mathbf{u}]$
		\State $\mathbf{V^{\perp}} \gets [\mathbf{V^{\perp}}, \mathbf{v}]$
		
		\EndFor
		\State return $\PrincipalAngles$
		
		\EndProcedure
	\end{algorithmic}
\end{algorithm}
\end{minipage}
\end{figure}

A more detailed analysis of these correlations can be obtained by computing the affinity between the subspaces spanned by each class of the MNIST dataset. For successful clustering, it is desired that different classes belong to dissimilar subspaces. The affinity between two $D$-dimensional subspaces can be measured by computing the principal angles 
$\PrincipalAngles = [\PrincipalAngle{1}, \PrincipalAngle{2}, ..., \PrincipalAngle{D}]^T ; \ \PrincipalAngle{i} \, \in \, [0,\pi/2] $ for $ i \in \{1,2,...,D\}$, which correspond to the minimized angles between two subspaces~\cite{Bjorck_AnglesBetweenLinearSubspaces_1973}. Angles close to 0 imply that vectors defining the subspaces are nearly colinear, whereas values close to $\pi / 2 $ correspond to almost orthogonal basis and
lead to smaller classification and clustering errors~\cite{Huang_RoleOfPrincipalAngelsSubspaceClassification_2015}.
\Algorithm{algorithm:principal angles algorithm} illustrates the greedy iterative method used to compute the principal angles between two subspaces.

\begin{table}[tb]
	\centering
	\captionof{table}{Principal angles between subspaces spanned by MNIST classes containing digits 0 and 2 in the image and scattering domains. Subspaces in the scattering domain are much more similar.}
	\label{table:principal angles}
	\begin{tabular}{|lr|ccccc|}
		\hline
		&&  \multicolumn{5}{c|}{Principal Angles (degrees)}\\
		Domain && $\PrincipalAngle{1}$ & $\PrincipalAngle{2}$ & $\PrincipalAngle{3}$ & $\PrincipalAngle{4}$ & $\PrincipalAngle{5}$ \\
		\hline
		
		Image && 63.0\degree & 63.8\degree & 65.5\degree & 68.7\degree & 69.5\degree  \\
		
		Scattering && 36.2\degree & 36.9\degree & 44.2\degree & 60.9\degree & 66.5\degree  \\
		
		\hline
		
	\end{tabular}
\end{table}

To measure the correlations between scattering coefficients, we compute the principal angles between the eigenbasis of two MNIST classes. For illustrative purposes, we select the classes containing digits 0 and 2 respectively.
Top five principal angles between those two classes are listed in \Table{table:principal angles}.
Whereas in the image domain all angles have magnitudes larger that $60\degree$, three principal angles have much smaller magnitude in the scattering domain, indicating a high correlation between the eigenbasis of two different classes. The highly correlated eigenvectors are precisely the ones corresponding to the largest eigenvalues.

High correlations and small principal angles between different classes are undesirable properties for the task of clustering in particular, and for pattern recognition tasks in general \cite{Huang_RoleOfPrincipalAngelsSubspaceClassification_2015}. The correlations might mislead the clustering algorithm resulting in errors. In the following section, we introduce a simple, but effective, algorithm to remove the correlations introduced by the scattering transform.

\subsection{Projection onto Orthogonal Complement}\label{sec:Projection onto Orthogonal Complement}

Motivated by the strong correlations between the subspaces corresponding to different image classes in the scattering domain, we propose the POC algorithm to improve the scattering representation for the task of clustering.

The POC algorithm applies a linear operator that removes the high correlations between the subspaces corresponding to  different classes, while preserving the specificity of the scattering features. This is achieved through an orthogonal projection that removes the principal scattering directions of variance.

For our image clustering pipeline, the POC algorithm receives as input the scattering transform of an image dataset in matrix form $\ScatteringFeatures \in \mathbb{R}^{D \times N}$. Each column of this matrix corresponds to the $D$-dimensional scattering transform of one image from the dataset.

First, we compute the Karhunen-Loeve transform of the scattering matrix, thus yielding the eigenvalues $\Eigenvalues = [\Eigenvalue{1}, \Eigenvalue{2}, \ldots, \Eigenvalue{D}]$ and eigenvectors $\Eigenvectors = [\Eigenvector{1}, \Eigenvector{2}, \ldots, \Eigenvector{D}] \in \mathbb{R}^{D \times D}$. 

Second, we define an orthogonal projection matrix $\Eigenvectors' = [\Eigenvector{n+1}, \Eigenvector{n+2}, \ldots, \Eigenvector{D}] \in \mathbb{R}^{D \times (D-n)}$  by removing the eigenvectors corresponding to the $n$ largest eigenvalues. As explained in \Section{sec:Geometry of Scattering Transforms}, these eigenvectors are highly-correlated between different classes. 

Finally, we use $\Eigenvectors'$ to perform an orthogonal projection of the scattering matrix into a lower-dimensional space where scattering features with largest intra-class variations have been removed
\begin{align}
P(\ScatteringFeatures) = (\Eigenvectors')^T  \ScatteringFeatures.
\end{align}

To simplify the notation, we refer to $P(\ScatteringFeatures)$ as $\DatasetSamples$.

\begin{figure*}[bt]	
	\begin{subfigure}{0.33\textwidth}	
		\includegraphics[height=0.2\textheight]{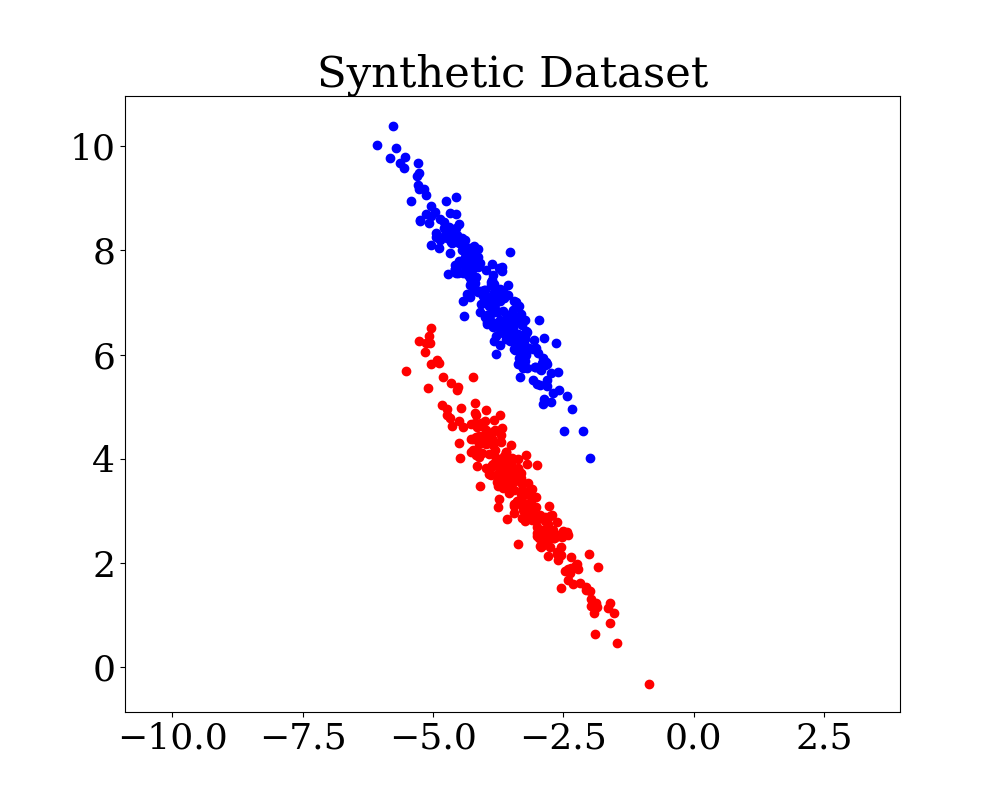}
		\label{fig:subfig1}
	\end{subfigure}
	\begin{subfigure}{0.33\textwidth}
		\centering
		\includegraphics[height=0.2\textheight]{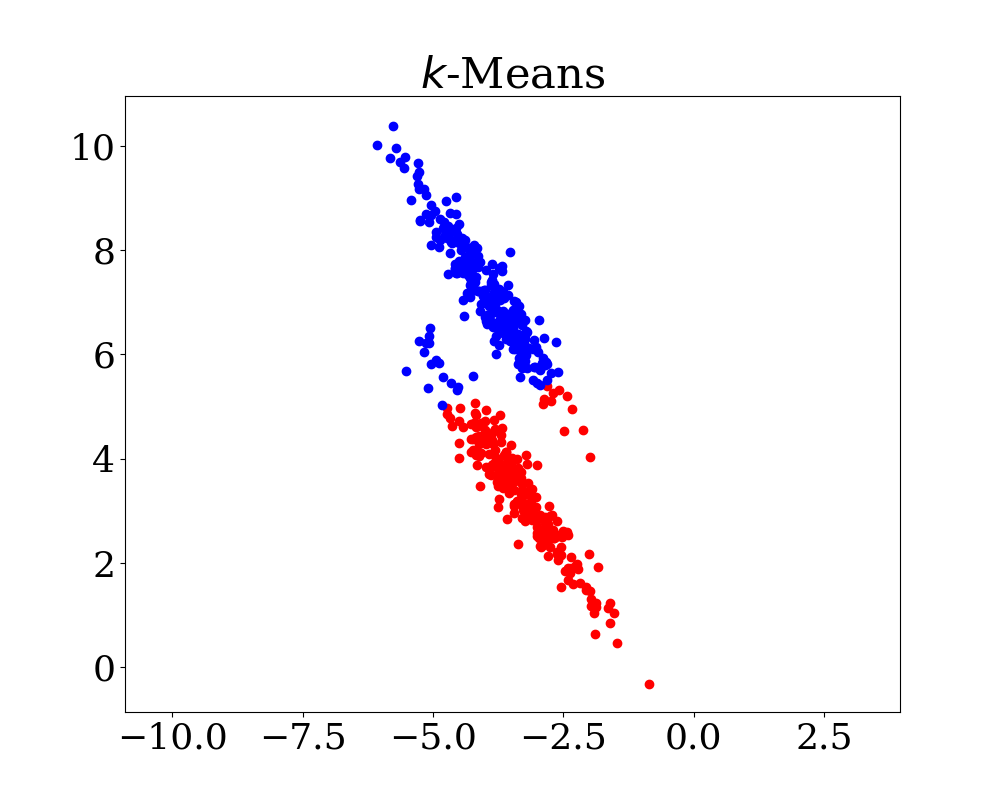}
		\label{fig:subfig2}
	\end{subfigure}
	\begin{subfigure}{0.33\textwidth}
		\includegraphics[height=0.2\textheight]{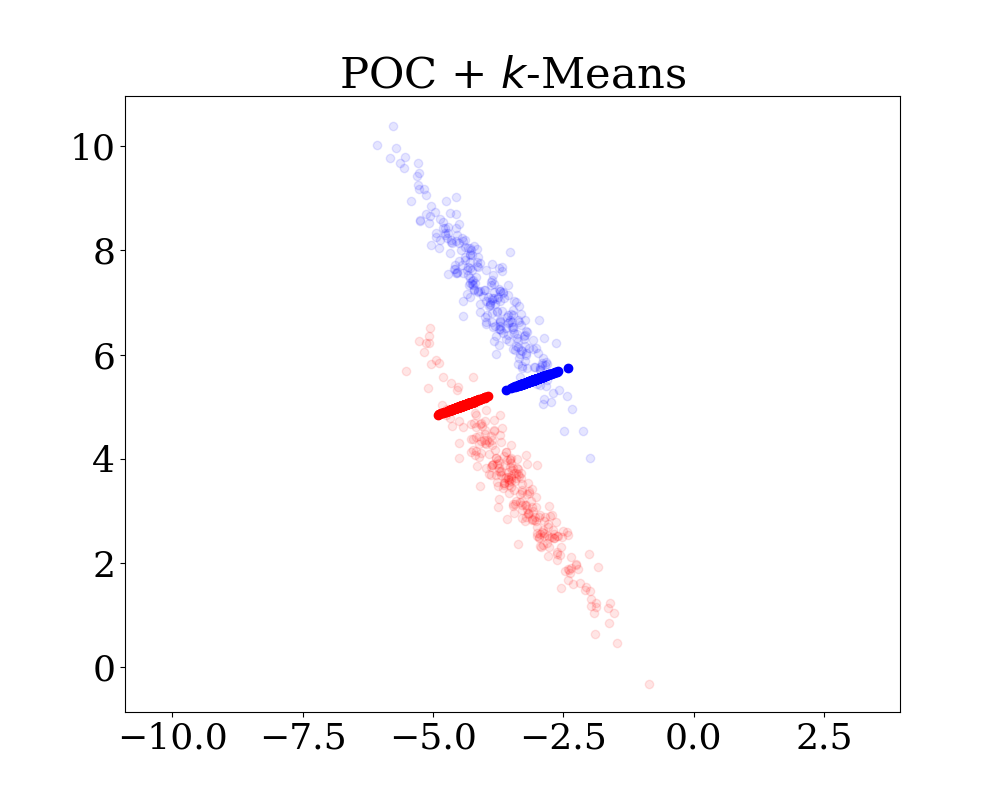}
		\label{fig:subfig1}
	\end{subfigure}
	\vspace{-0.3cm}
	\caption{Illustration of the POC step for clustering. Left-hand side figure displays a toy dataset composed of two elongated clusters. The figure in the center displays the predictions obtained using $k$-means. Right-hand side figure displays the projected samples and cluster assignments obtained applying POC followed by $k$-means. It is shown how the POC step reduces intra-class variabilities, simplifying the task of clustering.}
	\label{fig:illustrating_poc}
\end{figure*}

\Figure{fig:illustrating_poc} illustrates the effect of the POC algorithm for the task of clustering. For illustrative purposes, we use a toy dataset composed of two elongated clusters, displayed on the left-hand side. The shape of the clusters imitates the geometric structure of the scattering transform of small images, in which few directions span most of the data variance.

The figure in the center displays the cluster predictions obtained using $k$-means. Large intra-class variability leads the clustering algorithm to assign several samples to the wrong cluster.

The figure on the right hand side displays the projected samples and cluster assignments obtained applying POC followed by $k$-means. 
The POC projection step removes the direction that spans the largest variance, which is shared among the two classes and leads to errors when directly applying the clustering algorithm. 
The correct cluster assignments are obtained applying $k$-means to the projected data, once the intra-class variabilities are reduced.

Despite the simplicity of this method, applying POC substantially increases the image clustering performance of PSSC framework.

The POC algorithm shares similarities with the PCA dimensionality reduction algorithm. However, instead of dropping the eigenvectors corresponding to the \textit{smallest} eigenvalues, the geometry of the scattering transform and the high correlations between scattering coefficients motivate us to remove the eigenvectors corresponding to the \textit{largest} eigenvalues.

\subsection{Clustering Approach}\label{sec:Clustering Approach}

The final step in our image clustering pipeline, as depicted in \Figure{fig:clustering pipeline}, is a spectral clustering algorithm, which groups the $N$ processed feature vectors into $\NumClusters$ disjoint clusters.
In particular, we use a modified version of ultra-scalable spectral clustering (U-SPEC)~\cite{Huang_UltraScalableSpectralClustering_2019}. This method
allows to deal with non-linearly separable clusters, while scaling efficiently to large datasets.

U-SPEC scales gently with the size of the dataset through the use of two relaxations: a hybrid sampling approach to select a robust subset of $\Representatives$ representative samples from the dataset, and an efficient approximate $k$-NN search to construct a sparse affinity sub-matrix between the $N$ samples from the dataset and the selected $\Representatives$ representatives.

First, instead of computing an affinity matrix using all samples from the dataset, a robust subset of $\Representatives$ examples is sampled using a hybrid approach. Closely following \cite{Huang_UltraScalableSpectralClustering_2019}, a set of $\Candidates$ ($\Representatives < \Candidates \ll N$) candidate data points is randomly sampled from the dataset. Then, we perform the $k$-means algorithm on the $\Candidates$ candidates to obtain $\Representatives$ clusters, and use the $\Representatives$ cluster centers as the set of representatives.

Second, we compute a sparse affinity sub-matrix  between the $N$ samples from the dataset and the selected $\Representatives$ representatives using an approximate $k$-nearest neighbor ($k$-ANN) algorithm. This affinity sub-matrix $\AffinityMatrix \in \mathbb{R}^{N \times \Representatives}$ encodes pairwise relationships between the representatives
$\RepresentativesVector = [\mathbf{\Representatives}_1,  \mathbf{\Representatives}_2, ...,  \mathbf{\Representatives}_{\Representatives}]$ and the samples from the dataset 
$\DatasetSamples = [\Sample{1}, \Sample{2}, ..., \Sample{N}]$.
More precisely, each row $i=1,...,N$ will contain $k$ non-zero entries corresponding to the $k$-NN of $\Sample{i}$:
\begin{align}
& \AffinityMatrixEntry{i}{j} = 
\begin{cases}	
\exp \left(-\frac{||\mathbf{s}_i - \mathbf{p}_j||^2_2}{2\sigma^2} \right), &\mathbf{\Representatives}_j \text{ in } k\text{-neighborhood of } \Sample{i} \quad 
\\
0 , & \text{otherwise.}
\end{cases}
\end{align}

The affinity sub-matrix entries contain only $k \cdot N$ non-zero entries out of the total $\Representatives \cdot N$ ($k \ll \Representatives$), hence making it a memory-efficient representation.

Differently from \cite{Huang_UltraScalableSpectralClustering_2019}, we use Hierarchical Navigable Small World (HNSW) graphs \cite{Malkov_HNSWG_2018} to perform an efficient and robust $k$-ANN search for the construction of the affinity sub-matrix. This method exploits hierarchical directed graphs to find approximate nearest neighbors. We follow this approach to compute the affinities since, to the best of our knowledge, it outperforms all other $k$-ANN methods for high dimensional datasets, such as the scattering transform of small images.

Finally, U-SPEC aims at obtaining cluster assignments by treating the \emph{Laplacian} of the sparse affinity sub-matrix $\AffinityMatrix$ as a graph and solving an eigenproblem~\cite{Luxburg_SpectralClusteringTutorial_2007}. However, directly solving the eigenproblem is a computationally expensive step that takes $O(N^3)$.

Li \EtAl.~\cite{Li_SegmentationUsingSuperpixelsTransferCut_2012} show how this step can be solved more efficiently by exploiting the bipartite structure of $\AffinityMatrix$. Applying \emph{transfer cut}~\cite{Li_SegmentationUsingSuperpixelsTransferCut_2012}, the clustering is reduced to solving the generalized eigenproblem \eqref{eq:eigenproblem} using the Laplacian $\Laplacian$ of a much smaller graph $\WMatrix$, with a much lower complexity of $O(\Representatives^3)~(\Representatives \ll N)$:

\vspace{-0.2cm}
\begin{align}
& \WMatrix = \AffinityMatrix^T \DegreeMatrix_X^{-1} \AffinityMatrix   \label{eq:reduce_graph} \\
& \Laplacian = \DegreeMatrix_Y^{-1/2} \WMatrix \DegreeMatrix_Y^{-1/2} \label{eq:laplacian} \\
& \Laplacian \Eigenvector{} = \lambda \DegreeMatrix_Y \Eigenvector{} \label{eq:eigenproblem}.     
\end{align}

Above, $\DegreeMatrix_X$ and $\DegreeMatrix_Y$ are the degree matrices of $\AffinityMatrix$. These are diagonal matrices in which each element $(i,i)$ from the diagonal is equal to the sum of $\AffinityMatrix$'s $i_{\text{th}}$ row or column respectively. 

After having solved the eigenproblem \eqref{eq:eigenproblem}, the eigenvectors associated to the $\NumClusters$ largest eigenvalues are stacked into a matrix $\ClusteringMatrix = [\Eigenvector{1}, ..., \Eigenvector{\NumClusters}] \in \mathbb{R}^{N \times \NumClusters}$.

Finally, the cluster assignments are obtained by treating each row of $\ClusteringMatrix$ as a feature vector and clustering them using $k$-means. An original sample $\Sample{i}$ will be assigned to cluster $c$ if and only if the $i_{\text{th}}$ row of $\ClusteringMatrix$ was assigned to cluster $c$.

\section{Experimental Setup}\label{sec:experimental setup}

In this section, we describe the image datasets used for testing, implementation details, the different evaluation metrics for the task of image clustering and the baseline methods used for comparison.

\subsection{Datasets}

We evaluate our image clustering pipeline and compare it with other popular clustering methods on four different benchmark datasets. The MNIST \cite{LeCun_MnistDataset_1998} database consists of 70000 images of size $28 \times 28$ pixels each of handwritten digits from ten different categories (digits 0--9). Furthermore, we also consider the MNIST-test set, which contains a subset of 10000 images from MNIST. The USPS dataset \cite{Hull_USPSDataset_1994} consists of 9298 handwritten digit images of size $16 \times 16$ each. Fashion-MNIST \cite{Xiao_FashionMNIST_2017} consists of 70000 images of size $28 \times 28$ pixels each of fashion products divided into 10 categories. This dataset poses a significantly more challenging alternative to the MNIST dataset due to large variability of the images in each category.

\Table{table:image clustering datasets} displays an overview of the datasets used for evaluation of our image clustering pipeline.

\begin{table}
	\centering
	\captionof{table}{Image datasets used for evaluting the image clustering pipeline.}
	\label{table:image clustering datasets}
	\begin{tabular}{|llll|}
		\hline
		Dataset &  Images &  Classes & Image Size \\
		\hline
		MNIST \cite{LeCun_MnistDataset_1998} & 70000 & 10 & $28 \times 28$ \\
		MNIST-test \cite{LeCun_MnistDataset_1998} & 10000 & 10 & $28 \times 28$ \\
		USPS \cite{Hull_USPSDataset_1994} & 9298 & 10 & $16 \times 16$ \\
		Fashion-MNIST \cite{Xiao_FashionMNIST_2017} & 70000 & 10 & $28 \times 28$  \\
		\hline
		
	\end{tabular}
\end{table}

\subsection{Evaluation Metrics}

To evaluate the performance of the image clustering methods, we use two metrics widely used in the unsupervised learning literature. Both metrics are in the range [0,1], with higher values indicating a better clustering performance. Let $N$ denote the total number of samples, $\mathbf{y}$ denote to the ground truth labels and $\mathbf{\hat{y}}$ denote the predicted cluster assignments.

\begin{itemize}
	\item Clustering Accuracy (ACC) is the best match between predicted clustering assignments and ground truth:
	
	\vspace{-0.3cm}
	\begin{align}
	& ACC = \max_c \frac{\sum_{i=1}^{N} \mathbf{1}({y_i = \hat{y}_i^c})} {N}
	\end{align}
	
	where $\mathbf{\hat{y}}^c$ is the $c$th permutation of $\mathbf{\hat{y}}$.
	
	\item Normalized Mutual Information \cite{Vinh_InformationTheoreticMeasuresClusteringEvaluation_2010} (NMI) is a metric inspired by information theory, which corresponds to a normalization of the mutual information between the predicted cluster assignments and the ground truth labels. This metric is symmetric and invariant to label permutations. NMI is defined as:
	
	\vspace{-0.4cm}
	\begin{align}
	& NMI = \frac{2 \cdot I(\mathbf{y}, \mathbf{\hat{y}})}{H(\mathbf{y}) + H(\mathbf{\hat{y}})}
	\end{align}
	
	where $I(\cdot, \cdot)$ is the mutual information and $H(\cdot)$ the binary entropy.
	
\end{itemize}

\begin{table*}[t]
	\centering
	\caption{Results of the compared methods for the benchmark datasets. Methods above the double-line are shallow clustering methods, whereas the ones below are deep clustering methods. Best results from each category are highlighted in bold. Exact results were taken from the literature or computed by us. Results marked with `--' indicate out of memory error.}
	\label{table:clustering evaluation}
	\scriptsize
	\begin{adjustbox}{width=0.99\textwidth}
		\begin{tabular}{lr|llrllrllrll}
			\hline
			&&  \multicolumn{2}{c}{MNIST} && \multicolumn{2}{c}{MNIST-test} && \multicolumn{2}{c}{USPS} && \multicolumn{2}{c}{Fashion-MNIST} \\
			
			Method && ACC & NMI && ACC & NMI  && ACC & NMI  && ACC & NMI \\
			
			\hline
			
			$k$-means~\cite{Macqueen_KMeansClustering_1967}
			&& $0.584 $  & $0.497 $ && $0.596$ &  $0.505$
			&& $0.735 $ &  $0.613 $ && $0.474$ &  $0.512$ \\
			
			SC~\cite{Ng_SpectralClustering_2002}
			&& -- & -- && $0.705 $ & $0.715 $ 
			&& $0.818 $ & $0.819  $ && -- & -- \\
			
			DBSCAN~\cite{Ester_DBSCANDensityBasedClustering_1996}
			&& -- & -- && $0.114 $ & $0.167 $ 
			&& $0.000 $ & $0.000 $ && -- & -- \\
			
			PSSC (ours) && $\mathbf{0.965} $ & $\mathbf{0.913}$ && $\mathbf{0.967}$ & $\mathbf{0.919}$
			&& $\mathbf{0.957}$ & $\mathbf{0.898}$ && $\mathbf{0.628}$ & $\mathbf{0.644}$ \\
			
			\hline
			\hline
			
			DEC~\cite{Xie_UnsupervisedDeepEmbeddingClustering_2016}
			&& 0.863 & 0.834 && 0.856 & 0.830 
			&& 0.762 & 0.767 && 0.518 & 0.546 \\
			
			IDEC~\cite{Guo_ImprovedDeepEmbeddedClustering_2017}
			&& 0.881& 0.867&& 0.846& 0.802
			&& 0.761& 0.758&& 0.529& 0.557\\
			
			JULE~\cite{Yang_JointUnsupervisedLearningImageClusters_2016joint}
			&& 0.964& 0.913&& 0.961& 0.915
			&& 0.950& 0.913&& 0.563& 0.608\\
			
			DCN~\cite{Yang_TowrdsKMeansFriendlySpacesDeepClustering_2017}
			&& 0.830 & 0.810 && 0.802 & 0.786 
			&& 0.688 & 0.683 && 0.501 & 0.558 \\
			
			DEPICT~\cite{Ghasedi_DEPICTDeepClusteringJointConvolutionalAutoencoderAndRelativeEntropyMinimization_2017} 	
			&& 0.965& 0.917&& 0.963& 0.915
			&& 0.964& 0.927&& 0.392& 0.392\\

			DDC~\cite{Ren_DeepDensityBasedClustering_2020}
			&& 0.965& 0.932&& 0.965& 0.916
			&& 0.967& 0.918&& 0.619& 0.682\\
			
			N2D~\cite{Mcconville_DeepClusteringManifoldAutoencodedEmbedding_2019}
			&& 0.979& 0.942&& 0.948& 0.882
			&& 0.958& 0.901&& $\mathbf{0.672}$& $\mathbf{0.684}$\\
			
			ADEC~\cite{Mrabah_AdversarialDeepEmbeddedClustering_2020}
			&& $\mathbf{0.986}$& $\mathbf{0.961}$&& $\mathbf{0.985}$& $\mathbf{0.957}$
			&& $\mathbf{0.981}$& $\mathbf{0.948}$&& 0.586& 0.662\\
			
			\hline
		\end{tabular}
	\end{adjustbox}
\end{table*}

\subsection{Implementation Details}

Our methods are implemented using the Python programming language, using the packages Numpy~\cite{Harris_Numpy_2020} and Scikit-Learn~\cite{Pedregosa_ScikitLearn_201}. The scattering transforms are computed using the Kymatio software package~\cite{Andreux_Kimatio_2018}.

The scattering transforms are computed on a MSI Nvidia GeForce GTX 1080 Ti GPU and all other computations are performed on an
$\text{Intel}^{\text{\textregistered}}$ $\text{Xeon}^{\text{\textregistered}}$ Silver 4114 CPU.

For the scattering transform, we use complex Morlet filters as convolutional kernels. We set the spatial support of the wavelets to $2^J = 8$ pixels and the number of angles to $L=8$. We consider the ScatNets of two levels of depth as recommended in \cite{Bruna_InvariantScattering_2012}.

Input images are zero-padded to a $32 \times 32$ resolution (1024-dimensional representation). Applying the scattering transform with the aforementioned settings to these images yields $D=3472$ scattering coefficients, roughly resulting in a $3.4$ times dimensionality increase.


Prior to applying the POC algorithm, we apply PCA to reduce the dimensionality of the scattering feature vectors to $D=1000$. As shown in \Figure{fig:eigenvalues mnist dataset}, a large number of eigenvalues explain close to zero variance. Therefore, removing the dimensions associated to these eigenvalues significantly reduces computations while having negligible effect on the clustering results.

The number of directions removed by the POC algorithm is the same for all four evaluated datasets.
Namely, the orthogonal projection matrix $\Eigenvectors' = [\Eigenvector{n+1}, ..., \Eigenvector{D}]$ is constructed by removing the top two eigenvectors ($n$=2).

Regarding the hyper-parameters for the U-SPEC algorithm, we select the number of candidates and representatives as $\Candidates= 9000$ and $\Representatives=1000$, respectively. We compute the affinity sub-matrix by retrieving the five-nearest neighbors ($k = 5$) and finally assign the samples into $\NumClusters$ groups, where $\NumClusters$ is the ground-truth number of clusters.

\subsection{Baseline Methods}\label{section: baseline methods}

We compare our proposed PSSC pipeline with several popular image clustering baselines, belonging to different algorithmic classes. From the shallow clustering methods, we select $k$-means~\cite{Macqueen_KMeansClustering_1967}, spectral clustering (SC)~\cite{Ng_SpectralClustering_2002} and DBSCAN~\cite{Ester_DBSCANDensityBasedClustering_1996}. From the deep clustering methods, we compare against DEC~\cite{Xie_UnsupervisedDeepEmbeddingClustering_2016}, IDEC~\cite{Guo_ImprovedDeepEmbeddedClustering_2017}, JULE~\cite{Yang_JointUnsupervisedLearningImageClusters_2016joint}, 
DEPICT~\cite{Ghasedi_DEPICTDeepClusteringJointConvolutionalAutoencoderAndRelativeEntropyMinimization_2017},
DCN~\cite{Yang_TowrdsKMeansFriendlySpacesDeepClustering_2017},
DDC~\cite{Ren_DeepDensityBasedClustering_2020}, ADEC~\cite{Mrabah_AdversarialDeepEmbeddedClustering_2020} and 
N2D~\cite{Mcconville_DeepClusteringManifoldAutoencodedEmbedding_2019}

The evaluation results reported in the following section were extracted from the literature, or are the average results of five independent trials using the released code repository.

\section{Evaluation}\label{sec:evaluation}

\subsection{Clustering Results}

We evaluate our proposed method (PSSC) and compare it to the baseline methods introduced in \Section{section: baseline methods}. \Table{table:clustering evaluation} reports the image clustering results measured by clustering accuracy (ACC) and normalized mutual information (NMI). Methods above the double-line correspond to shallow clustering algorithms that do not rely on deep learning for feature extraction, whereas the ones below the double-line correspond to deep clustering algorithms. For each column, the best method from each algorithmic category is highlighted in bold.
From \Table{table:clustering evaluation} we can extract the following observations.

First, deep clustering methods outperform by a large margin all algorithms not based on deep learning, with the exception of PSSC. This is due to the fact that the embedding representations learned by the CNNs are able to capture more semantically meaningful features, which then allow to perform clustering successfully.

Second, our PSSC pipeline performs better than all shallow clustering methods on all datasets by a large margin. In fact, our proposed method performs comparably to state-of-the-art deep clustering methods, achieving top-3 performance for at least one evaluation metric in three out of four evaluated datasets (all except USPS).
This fact shows that the post-processed scattering features obtained by PSSC are equivalent to those learned by CNNs for the task of clustering. The advantage of our method is that it does not need to undergo a learning procedure.

\subsection{Timing Results}

In \Table{table:time comparison} we show the amount of time required to cluster the entire MNIST dataset using different clustering algorithms. Except for PSSC, we exclude all other shallow clustering algorithms from the comparison since they do not yield competitive results.

\begin{table}[h]
	\centering
	\captionof{table}{Time required to cluster the MNIST dataset for different clustering algorithms. Out method presents a much better running time than the competing algorithms.}
	\label{table:time comparison}
	\begin{tabular}{|l|ccccc|}
		\hline
		& PSSC & DEC & JULE & DEPICT & N2D \\
		\hline
		Time (s) & 182  & 613 & 12500 & 9561 & 1080 \\
		\hline
	\end{tabular}
\end{table}

From the results listed in \Table{table:time comparison}, we clearly see how PSSC is more efficient than all baseline methods. 
Whereas deep clustering algorithms need to undergo a training procedure to learn appropriate weights, all parameters from the PSSC pipeline are fixed, making it an efficient clustering method. Precisely, compared to clustering algorithms that achieve the same performance for the MNIST dataset as PSSC (e.g, JULE or DEPICT), our method reduces the execution time by more than one order of magnitude.

\subsection{Ablation Study}

Our PSSC clustering framework is composed of three well differentiated building blocks: ScatNet, POC and spectral clustering. In this section, we perform an ablation study to investigate the contribution of each of the aforementioned building blocks to the clustering accuracy. In particular, we evaluate different versions of our original PSSC pipeline on the MNIST dataset. These versions are constructed by either removing some of the original building blocks (i.e., ScatNet or POC processing) or by replacing the U-SPEC clustering algorithm by a simpler method: $k$-means.

\Table{table:ablation study} displays the results of the ablation study from which we extract the following observations.

First, clustering using scattering representations instead of images yields a large performance boost, improving the clustering accuracy from 75\% (2) and (4) to 96.5\% (1). 

Second, experiments using our POC algorithm along with the scattering transform demonstrate a superior clustering performance. When removing the POC processing step from our original pipeline, the clustering accuracy drops from 96.5\% (1) to 94.2\% (3). 

The effect of POC is more significant when we replace the U-SPEC clustering algorithm by $k$-means. The latter computes the cluster assignments based on the distance between the cluster center and the samples. Therefore, it is beneficial to have the samples closely scattered around the cluster center. As explained in \Section{sec:Projection onto Orthogonal Complement}, the POC algorithm removes the directions of largest variance of the scattering transform. This processing step significantly reduces redundant intra-class variabilities, thus bringing samples much closer to the cluster center, as shown in \Figure{fig:illustrating_poc}. From \Table{table:ablation study}, we see that POC processing improves the accuracy when using $k$-means from 60.1\% (6) to 83.8\% (5).

Finally, we see how using U-SPEC for the clustering step clearly outperforms $k$-means. This is an expected results since $k$-means is not able to deal with non-linearly separable clusters.

\begin{table}[tb]
	\centering
	\captionof{table}{Results of the ablation study on the PSSC clustering pipeline. We compare the original PSSC pipeline (1) with different simplified versions.}
	\label{table:ablation study}
	\begin{tabular}{|lr|ccc|}
		\hline
		&&  \multicolumn{3}{c|}{Results}\\
		Building Blocks && ACC & NMI & Time (s)\\
		\hline
		(1) ~Scat + POC + U-SPEC && 0.965 & 0.913 & 182 \\
		(2) ~POC + U-SPEC && 0.750 & 0.687 & 107 \\
		(3) ~Scat + U-SPEC && 0.942 & 0.881 & 121 \\
		(4) ~U-SPEC && 0.757 & 0.711 & 43 \\
		(5) ~Scat + POC + $k$-means && 0.838 & 0.718 & 173 \\
		(6) ~Scat + $k$-means && 0.601 & 0.538 & 187 \\

		
		\hline
		
	\end{tabular}
\end{table}

\section{Conclusion}\label{sec:conclusion}

In this paper we propose a novel image clustering framework, PSSC, based on post-processed scattering features extracted from the input images. Our method achieves comparable results to those of state-of-the-art deep learning-based methods, while being robust, mathematically interpretable and efficient.

Our PSSC framework obtains the cluster assignments cascading three steps. First, we process the images using a ScatNet, thus generating image features that are invariant to small translations and with linearized intra-class variabilities. Second, scattering features are processed by the POC algorithm, which exploits the distribution of scattering features to project the samples into a lower-dimensional manifold more suitable for clustering. Finally, the projected features are clustered using an efficient and scalable spectral clustering algorithm.

We compare PSSC with both traditional algorithms and deep clustering methods. Our experiments demonstrate that PSSC obtains the best performance among all shallow clustering algorithms for all evaluated benchmark datasets. Furthermore, we achieve comparable results to those of the latest deep clustering techniques, reaching top-3 performance in three out of four evaluated datasets while reducing the execution time by more than one order of magnitude. 

A natural next step is to address the image clustering problem for more complex datasets, such as
Imagenet. We think that the method developed in this paper may be used as a building block towards solving
this challenging problem. Since Imagenet is very large and contains a lot of small image patches, it is critical
to have fast and robust clustering algorithm. In this sense, the order of magnitude speedup, with only a modest performance loss, of our method compared to deep clustering algorithms is very significant.


\clearpage

\bibliographystyle{IEEEtran}
\bibliography{referencesAngel.bib}
\end{document}

%% file: commands.tex
\usepackage{algorithm}
\usepackage[noend]{algpseudocode}
\usepackage{gensymb}
\usepackage{xcolor}
\usepackage{adjustbox}
\usepackage{tabularx}
\usepackage{amsmath}
\usepackage{amsthm}
\usepackage{textcomp}

\makeatletter
\def\BState{\State\hskip-\ALG@thistlm}
\makeatother

\theoremstyle{definition}

\newcommand{\EtAl}{\textit{et al}}

\newcommand{\MotherWavelet}{\psi}
\newcommand{\ScalingFunction}{\phi}
\newcommand{\Orientation}{\gamma}

\newcommand{\RotationGroup}{G}
\newcommand{\Input}{X}
\newcommand{\ScatteringCoefficients}{S}
\newcommand{\Path}{p}

\newcommand{\ImageDataset}{\mathcal{X}}
\newcommand{\Image}{\mathbf{X}}

\newcommand{\ScatteringFeatures}{S_J\mathcal{X}}
\newcommand{\PrincipalAngles}{\boldsymbol{\theta}}
\newcommand{\PrincipalAngle}[1]{\theta_{#1}}

\newcommand{\Eigenvalues}{\boldsymbol{\lambda}}
\newcommand{\Eigenvalue}[1]{\lambda_{#1}}
\newcommand{\Eigenvectors}{\mathbf{W}}
\newcommand{\Eigenvector}[1]{\mathbf{w}_{#1}}

\newcommand{\Candidates}{p'}

\newcommand{\Representatives}{p}
\newcommand{\Sample}[1]{\mathbf{s}_{#1}}
\newcommand{\RepresentativesVector}{\mathbf{P}}
\newcommand{\DatasetSamples}{\mathbf{S}}
\newcommand{\AffinityMatrix}{\mathbf{A}}

\newcommand{\AffinityMatrixEntry}[2]{A_{{#1},{#2}}}
\newcommand{\DegreeMatrix}{\mathbf{D}}
\newcommand{\WMatrix}{\bar{\mathbf{A}}}
\newcommand{\Laplacian}{\mathbf{L}}
\newcommand{\NumClusters}{C}

\newcommand{\ClusteringMatrix}{\mathbf{C}}

\newcommand{\Table}[1]{Table~\ref{#1}}
\newcommand{\Figure}[1]{Figure~\ref{#1}}

\newcommand{\Algorithm}[1]{Algorithm~\ref{#1}}

\newcommand{\Section}[1]{Section~\ref{#1}}